\documentclass[11pt]{article}

\usepackage[utf8]{inputenc}
\usepackage[T1]{fontenc}
\usepackage[margin=1in]{geometry}
\usepackage{amsmath,amssymb}
\usepackage{natbib}
\setcitestyle{authoryear,round,semicolon}
\usepackage{booktabs}
\usepackage{fancyvrb}
\usepackage{graphicx}
\usepackage{pgfplots}
\pgfplotsset{compat=1.18}
\usetikzlibrary{arrows.meta,positioning,fit,backgrounds}
\usepackage{array}
\usepackage{hyperref}
\hypersetup{colorlinks=true,linkcolor=blue,citecolor=blue,urlcolor=blue}

\title{\bfseries On the Boundary of Admission Gates:\\ An Injected-Truth Study of Falsification-First\\ Selection in Quantitative Strategy Research}

\author{%
  Tianlun Zheng\thanks{Correspondence: \texttt{simplify23@users.noreply.github.com}. %
Code and data to reproduce every number in this paper are available at %
\url{https://github.com/simplify23/quant-trading-agent}.}\\[4pt]
  \small Fudan University%
}

\date{October 2026}

\begin{document}
\sloppy
\maketitle

\begin{abstract}
\noindent
Strategy research conflates two problems: finding a profitable rule, and establishing that
the finding is not search luck. The latter calls for \emph{admission gates}---statistical
criteria that must be satisfied before a conclusion is adopted---yet whether gates work, and
at what cost, remains untested. We introduce an \textbf{injected-truth protocol} with a
\textbf{random-admission control} that adopts at the same rate as the gate; only if the gate
beats this control does it carry information rather than merely raise a threshold. Across
synthetic and real-calibrated panels, gates eliminate false discoveries in the weak-signal
regime but cut adoption to $1$--$7\%$, and add nothing when signals are strong. Most
importantly, criteria computed on \emph{absolute} rather than \emph{excess} returns silently
reject every candidate, including true signals.

\vspace{4pt}
\noindent\textbf{Keywords:} multiple testing, backtest overfitting, strategy admission,
injected-truth validation, excess returns, false discovery rate
\end{abstract}

\section{Introduction}
\label{sec:intro}

\subsection{Two kinds of failure}

Data-driven strategy research fails in two qualitatively different ways. \textbf{Type I
failure} is not finding: search is inefficient and genuine signals remain undiscovered.
\textbf{Type II failure} is finding something false: after searching many candidates on a
finite sample, noise is mistaken for signal. The statistical root of Type II failure is
elementary---when one tests $N$ strategies and reports the best, the reported statistic is not
any single strategy's performance but an \emph{extreme value}---but its consequences are not.

\subsection{The engineering response, and what it assumes}

Practitioners and automated research systems respond by inserting \emph{admission gates}: a set
of criteria that a conclusion must satisfy before it is adopted. Current agentic pipelines
instantiate this as multi-stage funnels. \citet{factorMiner} report a $92\%$ elimination rate
across information-coefficient (IC) screening, correlation filtering, deduplication, and backtesting;
\citet{alphaagent} enforce originality, hypothesis--factor alignment, and complexity control;
\citet{minerva} compose four established validation statistics into a robustness grade. The
idea is natural. It is also largely untested, and two pitfalls follow.

\paragraph{Conservatism is not effectiveness.} Gates usually manifest as \emph{adopting less}.
Without a control one cannot distinguish a gate that carries information from one that simply
raised a threshold. A high elimination rate does not settle the question: random elimination at
the same rate might do equally well.

\paragraph{A gate can fail without announcing it.} \citet{minerva} calibrate their score on
$359{,}062$ production backtests, yet in their \emph{pre-registered} real-market test it showed
no significant forward relationship (Spearman $\rho_s=0.013$, $p=0.40$); the authors describe it
as ``an auditable validation and reporting layer, rather than evidence of demonstrated
real-market predictability.'' Having gates is not the same as gates being effective.

\subsection{Contributions}

We ask \emph{when} admission gates are useful and \emph{at what cost}, using an
\textbf{injected-truth protocol} in which known signals are embedded so that ground truth is
observable. Its essential component is a \textbf{random-admission control} that adopts at the
same rate as the gate but by coin flip, separating ``carries information'' from ``is more
conservative.'' Our contributions:

\begin{enumerate}\itemsep1pt
\item A reproducible evaluation protocol for admission gates, including the equal-adoption-rate
      control (\S\ref{sec:arms}).
\item An empirical boundary: gates help for training-period $t<2.5$ and are pure overhead
      beyond it; the cost is quantified in adoption rate and recall (\S\ref{sec:results}).
\item A necessary condition: criteria must be evaluated on \emph{excess} returns. On absolute
      returns they systematically and \emph{silently} reject every candidate, including a
      ground-truth signal with $t=4.0$ (\S\ref{sec:diag}).
\item Honest negatives: a redundant sample-size criterion, failed control designs, and a task
      morphology in which gates are unhelpful (\S\ref{sec:ablation}--\ref{sec:morphology}).
\end{enumerate}

\paragraph{Why it matters.} Every automated research pipeline that admits conclusions must
decide \emph{when to stop believing its own output}. Gates are the instrument that decision rests
on, yet their effectiveness is typically assumed rather than measured---and the one published
calibration of such an instrument reported a null forward result \citep{minerva}. If gates are
weaker than assumed, the pipelines that rely on them inherit an unbounded false-discovery risk;
if they are merely conservative, those pipelines silently discard real findings. Neither failure
is visible from the outside, which is what makes the question worth measuring.

We do not claim that gates increase returns, nor that our implementation improves on prior work.

\section{Related Work}
\label{sec:related}

Our paper sits at the intersection of four literatures, and belongs to none of them. We state
the branch explicitly in \S\ref{sec:position}.

\subsection{The replication crisis, quantified in finance}

\citet{ioannidis} argued that for most study designs the probability that a claimed finding is
true is below one half, driven by small samples, small effects, and the number of relationships
probed. Finance supplies unusually clean measurements of the same phenomenon.
\citet{mcleanpontiff} tracked 97 published cross-sectional predictors and found returns $26\%$
lower out-of-sample and $58\%$ lower post-publication. \citet{harveyliuzhu} showed that after
accounting for the number of factors tried, a newly claimed factor needs $t>3.0$, not $2.0$.
\citet{houxuezhang} re-ran 452 published anomalies: $65\%$ fail a single $|t|\ge1.96$ hurdle,
rising to $82\%$ under the multiple-testing hurdle of $2.78$. This literature establishes that
Type II failure is not hypothetical---it is the modal outcome in this field.

\subsection{Statistical machinery for data snooping}

The corrective tools are also mature. \citet{lomackinlay} documented data-snooping bias in
asset-pricing tests; \citet{brock1992} provided the canonical technical-rule study that later
work used to quantify snooping; \citet{stwbootstrap} evaluated trading rules by bootstrap
against the full universe they were drawn from. \citet{white2000} formalised the Reality Check
and \citet{hansen2005} refined it into the Superior Predictive Ability test. \citet{dsr}
introduced the deflated Sharpe ratio, and \citet{pbo} the probability of backtest overfitting;
\citet{harveyliu} showed the appropriate Sharpe deflation is non-linear in the strategy's own
statistics. We use the DSR \citep{dsr} directly as one of our four criteria.

\subsection{Robustness scoring and admission in practice}

A recent line composes these statistics into deployable instruments: \citet{minerva} aggregate
DSR, PBO, SPA and minimum track record length into a display score; \citet{gtscore} embeds
anti-overfitting terms directly into the optimisation objective. Both operate \emph{post hoc} on
a completed backtest. Our question is different in kind: not ``how robust is this result?'' but
``does the rule that decides whether to accept a result actually work, and what does it cost?''
The distinction matters because a scoring function's output is only as useful as the decision
rule that consumes it.

\subsection{Agentic alpha mining and automated research}

Automated pipelines attack Type I failure. \citet{tradingagents} simulate trading-firm
organisation with debating analyst agents; \citet{factorMiner} add a memory loop that distils
successful and failed mining attempts; \citet{quantaalpha} evolve whole mining trajectories;
\citet{alphaagent} constrain generation by abstract-syntax originality and hypothesis alignment,
and have been published at KDD~'25; \citet{automate} combine LLM factor generation with
multi-agent evaluation. \citet{alphaseek} pursue trajectory-level self-iteration. In the broader
automated-research literature, \citet{rsisurvey} survey 1{,}250 papers and observe that
demonstrated self-improvement tracks a verification hierarchy---formal verifiers $>$ execution
feedback $>$ learned judges $>$ intrinsic self-assessment---while \citet{aide2} close the loop on
an agent's own code. Our criteria sit at the \emph{execution feedback} level of that hierarchy.

\subsection{Where this paper sits}
\label{sec:position}

\begin{table}[htbp]
\centering
\caption{Positioning. This paper belongs to the methodology branch of the replication
literature, applied at admission time inside an automated research loop.}
\label{tab:position}
\small
\begin{tabular}{@{}p{2.5cm}p{3.0cm}p{2.9cm}p{3.0cm}p{3.2cm}@{}}
\toprule
& \textbf{Replication lit.} & \textbf{Robustness scoring} & \textbf{Agentic alpha mining} & \textbf{This paper} \\
\midrule
Question & Are published findings true? & How robust is this backtest? & How to find more alpha? & \textbf{Are the gates effective, and at what cost?} \\
Unit of analysis & Published factor & One completed backtest & Candidate factor & \textbf{A candidate \emph{decision}} \\
When applied & Post hoc, after publication & Post hoc, before deployment & At search time & \textbf{At admission time, inside the loop} \\
Output & Failure / FDR rates & A robustness grade & Factors and weights & \textbf{An effectiveness boundary} \\
Typical method & Surveys, replications & Composite statistics & LLM agents, evolution & \textbf{Injected truth $+$ random-admission control} \\
\bottomrule
\end{tabular}
\end{table}

Table~\ref{tab:position} states the positioning. Concretely, the paper belongs to the
\emph{methodology} branch of the replication literature (\S\ref{sec:related}.\!1--\ref{sec:related}.\!2):
it takes that branch's diagnosis as given, and instead of measuring how often findings fail,
it tests the instrument practitioners built in response. Relative to robustness scoring
(\S\ref{sec:related}.\!3) we change the object of study from a score to a decision rule, and
relative to agentic mining (\S\ref{sec:related}.\!4) our contribution is orthogonal and
composable: gates can serve as the admission layer of those pipelines, and a search-efficiency
improvement does not by itself make admission more reliable.

\section{Method}
\label{sec:method}

\subsection{Admission decision}

Let round $k$ propose a candidate $c_k$ with evidence $E_k$. The admission decision is
$\mathcal{A}(c_k,E_k)\to v_k\in\{\textsc{adopt},\textsc{pending},\textsc{reject},
\textsc{invalid}\}$, where \textsc{pending} denotes \emph{not detected} (insufficient power) and
is strictly distinguished from \textsc{reject} (evidence against). Our experiments use three
outcomes: \textbf{adopt}, \textbf{abstain}, and \textbf{false adoption}.

\subsection{Criteria}

We use four criteria (Table~\ref{tab:criteria}).

\begin{table}[htbp]
\centering
\caption{Admission criteria used in this study.}
\label{tab:criteria}
\small
\begin{tabular}{@{}p{3.0cm}p{8.2cm}p{2.6cm}@{}}
\toprule
\textbf{Criterion} & \textbf{Definition} & \textbf{Analogue} \\
\midrule
G6 sub-period stability & Four sub-periods of the training window; at least three have positive mean &
Parameter plateau \\
F6 leave-one-trade-out & After removing the best five trading days, the mean remains positive &
Single-event-carry check \\
F1 sample size & $n \geq N^{*}$, $\;N^{*}=\lceil (2\sigma/\mu)^2\rceil$ & Falsification threshold \\
G9a multiple testing & Deflated Sharpe ratio $\geq0.95$ given $N_{\text{trials}}=40$ & Trial-count deflation \citep{dsr} \\
\bottomrule
\end{tabular}
\end{table}

\subsection{A necessary condition: criteria on excess returns}
\label{sec:excess}

Write strategy $j$'s return on day $t$ as $r_{jt}=w_t+\varepsilon_{jt}$, where $w_t$ is the
benchmark return and $\varepsilon_{jt}$ the strategy-specific part. Total volatility is inflated
by the market component, $\sigma(r_j)\gtrsim\sigma(w)$, whereas true discernibility depends on
$\alpha/\sigma(\varepsilon)$, not on $\alpha/\sigma(r_j)$.

If criteria are evaluated on \emph{absolute} returns while market volatility dominates---in our
panels $\sigma(w)=0.0208$ versus $\sigma(\varepsilon)=0.007$---then the absolute Sharpe of
\emph{every} candidate, including the ground-truth one, lies inside the noise band and all four
criteria fail. The correction is to evaluate every criterion on the excess return
$x_{jt}=r_{jt}-w_t$; for the DSR this is the information ratio. We take $w_t$ to be the
equal-weighted panel return.

\subsection{Arms and controls}
\label{sec:arms}

Table~\ref{tab:arms} lists the four arms. Arm~C is the key control: with adoption rates aligned,
a criterion carries information if and only if
$\Delta\text{FDR}=\text{FDR}_{C}-\text{FDR}_{B}>0$. If $\Delta\text{FDR}\approx0$, the gate is
merely a stricter threshold.

\begin{table}[htbp]
\centering
\caption{Experimental arms. Arm C separates information from conservatism.}
\label{tab:arms}
\small
\begin{tabular}{@{}llp{7.8cm}@{}}
\toprule
\textbf{Arm} & \textbf{Rule} & \textbf{Purpose} \\
\midrule
A & Pure search & Adopt the candidate with the highest training-period Sharpe (current practice). \\
B & Gate & Among candidates passing all four criteria, adopt the highest training Sharpe; abstain if none pass (method under evaluation). \\
C & Random admission & Adopt arm A's choice with probability $p$, otherwise abstain; $p$ tuned so that C's adoption rate matches B's (identifies ``effective'' vs.\ ``merely conservative''). \\
B$'$ & Gate without F1 & Arm B with the sample-size criterion removed (ablation). \\
\bottomrule
\end{tabular}
\end{table}

\paragraph{Control design iterations.} Our first two designs for arm C were invalid. A
Sharpe-quantile threshold always leaves the training-period winner inside the passing set, so
``take the best passer'' is \emph{identical} to arm A---confirmed empirically, with identical
FDR at every $\alpha$. ``Randomly keep $k$, then take the best'' always adopts for $k\ge1$, so
its adoption rate is identically one and cannot be aligned. A valid control must be able to
\emph{abstain} \emph{and} have a tunable adoption rate.

\section{Experimental Setup}
\label{sec:setup}

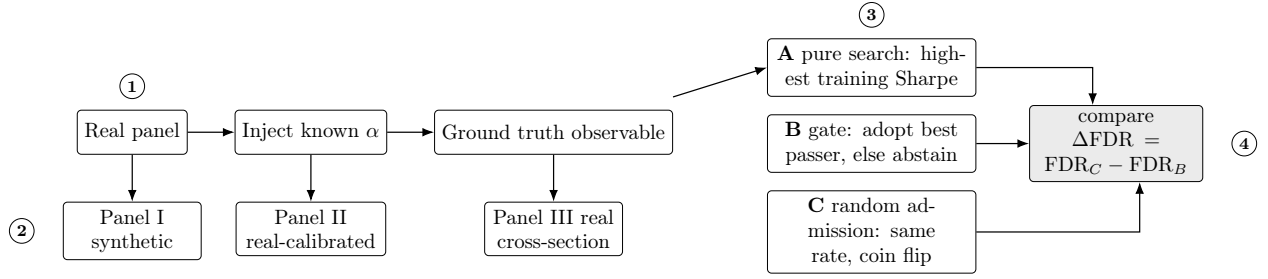
\begin{figure}[t]
\centering
\resizebox{\textwidth}{!}{%
\begin{tikzpicture}[
  font=\small,
  st/.style={circle, draw, inner sep=1.4pt, font=\scriptsize\bfseries, minimum size=4.2mm},
  box/.style={draw, rounded corners=2pt, align=center, inner sep=3.5pt, minimum height=7.5mm},
  arm/.style={draw, rounded corners=2pt, align=center, inner sep=3.5pt, minimum height=7mm},
  ar/.style={-{Latex[length=2mm]}, semithick}]
\node[box] (real) at (0,0) {Real panel};
\node[box, right=8mm of real] (inj) {Inject known $\alpha$};
\node[box, right=8mm of inj] (truth) {Ground truth observable};
\draw[ar] (real) -- (inj);
\draw[ar] (inj) -- (truth);
\node[st] at ([yshift=4mm]real.north) {1};
\node[box, below=8mm of real,  text width=21mm] (p1) {Panel I synthetic};
\node[box, below=8mm of inj,   text width=23mm] (p2) {Panel II real-calibrated};
\node[box, below=8mm of truth, text width=21mm] (p3) {Panel III real cross-section};
\draw[ar] (real.south) -- (p1.north);
\draw[ar] (inj.south) -- (p2.north);
\draw[ar] (truth.south) -- (p3.north);
\node[st] at ([xshift=-7mm]p1.west) {2};
\node[arm, right=16mm of truth, yshift=11mm, text width=33mm] (a) {\textbf{A}~pure search: highest training Sharpe};
\node[arm, below=3mm of a, text width=33mm] (b) {\textbf{B}~gate: adopt best passer, else abstain};
\node[arm, below=3mm of b, text width=33mm] (c) {\textbf{C}~random admission: same rate, coin flip};
\draw[ar] ([yshift=6mm]truth.east) -- (a.west);
\node[st] at ([yshift=4mm]a.north) {3};
\node[box, right=9mm of b, text width=27mm, fill=black!8] (cmp)
  {compare $\Delta\text{FDR}=\text{FDR}_C-\text{FDR}_B$};
\draw[ar] (a.east) -| ([xshift=-4mm]cmp.north);
\draw[ar] (b.east) -- (cmp.west);
\draw[ar] (c.east) -| ([xshift=4mm]cmp.south);
\node[st] at ([xshift=7mm]cmp.east) {4};
\end{tikzpicture}}
\caption{The injected-truth protocol, in four numbered steps: (1)~inject a known effect so that
ground truth becomes observable; (2)~instantiate it in one of three panel families; (3)~apply
three arms to the same candidate set---arm~C adopts at the \emph{same rate} as arm~B but by coin
flip; (4)~compare. Only if $\Delta\text{FDR}>0$ does the gate carry information rather than
merely being more conservative.}
\label{fig:protocol}
\end{figure}

Figure~\ref{fig:protocol} summarises the protocol.

\subsection{Data and power pre-check}

We use a panel of 29 A-share ETFs (745 return points, backward-adjusted); Table~\ref{tab:data}
reports its statistics. Before designing any portfolio-level comparison we compute the smallest
detectable effect. A leg with a 30-day minimum holding period yields at most $10$--$13$ trades
over the 322-day training window, whereas $N^{*}$ is $93$ trades for $\mu=1.25\%,\sigma=6\%$ and
$576$ for $\mu=0.5\%$. Portfolio-level comparison is therefore \emph{undecidable}
here---resolving it would require roughly 37 years of data---which is precisely why the
injected-truth design is needed (Appendix~\ref{app:repro} lists the reproduction commands).

\begin{table}[htbp]
\centering
\caption{Real panel statistics and power pre-check.}
\label{tab:data}
\small
\begin{tabular}{@{}lrr@{}}
\toprule
\textbf{Panel statistic} & \textbf{Value} & \textbf{Power item} \\
\midrule
Daily standard deviation & $0.02075$ & Training window: 322 days \\
Kurtosis (normal $=3$) & $7.84$ & Trades per leg (upper bound): 10--13 \\
First-order autocorrelation & $+0.021$ & $N^{*}$ ($\mu=1.25\%$): 93 trades \\
\textbf{Mean cross-sectional correlation} & $\mathbf{+0.504}$ & $N^{*}$ ($\mu=0.5\%$): 576 trades \\
\bottomrule
\end{tabular}
\end{table}

\subsection{Panels}

\paragraph{Panel I (synthetic, independent candidates).} $T=750$ (training 322), $N=40$
candidates of which $K=5$ are ground truth. Noise $\mathcal{N}(0,\sigma^2)$, $\sigma=0.01$;
ground-truth columns receive a constant $\alpha\in\{0.0004,0.0008,0.0016,0.0032\}$ (annualized
$10.1\%$--$80.6\%$). 400 repetitions per setting; $N\in\{20,40,80\}$ for sensitivity.

\paragraph{Panel II (real-calibrated strategy family).} The benchmark series is the equal-weighted
return of the real panel, preserving fat tails and volatility clustering; strategy-specific noise
$\sigma_\varepsilon\in\{0.007,0.004\}$ implies inter-strategy correlations of roughly $0.89$ and
$0.96$. Ground-truth columns receive $\alpha$ calibrated \emph{relative to $\sigma_\varepsilon$}
so that the intended $t$-statistics are actually attained.

\paragraph{Panel III (real cross-section, counter-example).} The 29 real ETFs are themselves
treated as candidates, i.e.\ an \emph{asset-selection} task (\S\ref{sec:morphology}).

\noindent Metrics: $\text{FDR}=$ share of false among adopted; \textbf{hit rate} $=$ share of runs
adopting a ground-truth signal; \textbf{abstention rate} $=1-$ adoption rate.

\section{Results and Analysis}
\label{sec:results}

\subsection{Panel I: help when signals are weak, overhead when they are strong}

\begin{table}[htbp]
\centering
\caption{Panel I (synthetic, independent candidates), 400 repetitions.}
\label{tab:exp1}
\small
\begin{tabular}{@{}lrrrrrrr@{}}
\toprule
$\alpha$ (ann.) & train $t$ & \textbf{A FDR} & B adopt & \textbf{B FDR} & \textbf{C FDR} & $\Delta$FDR & Verdict \\
\midrule
$10.1\%$ & 0.72 & $\mathbf{0.590}$ & 0.017 & $\mathbf{0.400}$ & 1.000 & $\mathbf{+0.600}$ & effective \\
$20.2\%$ & 1.44 & $\mathbf{0.243}$ & 0.047 & $\mathbf{0.000}$ & 0.200 & $\mathbf{+0.200}$ & effective \\
$40.3\%$ & 2.87 & 0.010 & 0.603 & 0.000 & 0.010 & $+0.010$ & \emph{indistinguishable} \\
$80.6\%$ & 5.74 & 0.000 & 1.000 & 0.000 & 0.000 & $+0.000$ & \emph{indistinguishable} \\
\bottomrule
\end{tabular}
\end{table}

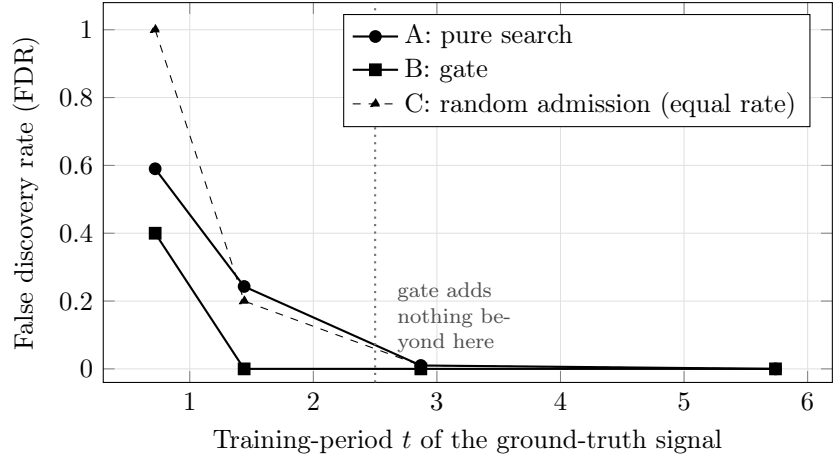
\begin{figure}[t]
\centering
\begin{tikzpicture}
\begin{axis}[
  width=0.68\textwidth, height=0.40\textwidth,
  xlabel={Training-period $t$ of the ground-truth signal},
  ylabel={False discovery rate (FDR)},
  xmin=0.3, xmax=6.2, ymin=-0.04, ymax=1.08,
  legend pos=north east, legend cell align=left,
  grid=both, grid style={line width=.2pt, draw=black!12},
  tick label style={font=\small}, label style={font=\small}, legend style={font=\small}]
\addplot[mark=*, thick, black] coordinates {(0.72,0.590) (1.44,0.243) (2.87,0.010) (5.74,0.000)};
\addlegendentry{A: pure search}
\addplot[mark=square*, thick, black] coordinates {(0.72,0.400) (1.44,0.000) (2.87,0.000) (5.74,0.000)};
\addlegendentry{B: gate}
\addplot[mark=triangle*, dashed, black] coordinates {(0.72,1.000) (1.44,0.200) (2.87,0.010) (5.74,0.000)};
\addlegendentry{C: random admission (equal rate)}
\draw[dotted, thick, black!55] (axis cs:2.5,-0.04) -- (axis cs:2.5,1.08);
\node[font=\scriptsize, anchor=south west, black!70, align=left, text width=20mm] at (axis cs:2.6,0.02) {gate adds nothing beyond here};
\end{axis}
\end{tikzpicture}
\caption{Panel~I (400 repetitions). Pure search errs on $24\%$--$59\%$ of weak-signal adoptions;
the gate drives FDR to zero but only below $t\approx2.5$. To the right of the dotted line the
gate is pure overhead: pure search barely errs and gating only lowers the adoption rate
(Table~\ref{tab:exp1}).}
\label{fig:fdr}
\end{figure}

Figure~\ref{fig:fdr} plots the same result as a curve:  In the weak regime pure search's FDR is
$24\%$--$59\%$; gates reduce it to zero at the cost of an adoption rate of $1.7\%$--$4.7\%$. For
training $t>2.5$, pure search already errs essentially never (FDR $\le1\%$) and gates add no
value while lowering the adoption rate.

\subsection{Panel II: stable across data source and correlation structure}

\begin{table}[htbp]
\centering
\caption{Panel II (real-calibrated strategy family), 400 repetitions each.}
\label{tab:exp2}
\small
\begin{tabular}{@{}lrrrrrr@{}}
\toprule
$\sigma_\varepsilon$ (corr.) & $t$ & \textbf{A FDR} & B adopt & \textbf{B FDR} & B hit & B abstain \\
\midrule
0.007 ($\approx0.89$) & 1.00 & $\mathbf{0.388}$ & 0.012 & $\mathbf{0.000}$ & 0.013 & 0.988 \\
0.007 & 1.51 & $\mathbf{0.188}$ & 0.073 & $\mathbf{0.034}$ & 0.070 & 0.927 \\
0.007 & 2.51 & 0.020 & 0.387 & $\mathbf{0.000}$ & 0.388 & 0.613 \\
0.007 & 4.00 & 0.003 & 0.982 & $\mathbf{0.000}$ & 0.983 & 0.018 \\
\midrule
0.004 ($\approx0.96$) & 0.99 & $\mathbf{0.415}$ & 0.017 & $\mathbf{0.000}$ & 0.018 & 0.983 \\
0.004 & 1.53 & $\mathbf{0.182}$ & 0.068 & $\mathbf{0.000}$ & 0.068 & 0.932 \\
0.004 & 2.51 & 0.018 & 0.395 & $\mathbf{0.000}$ & 0.395 & 0.605 \\
0.004 & 3.99 & 0.000 & 0.987 & $\mathbf{0.000}$ & 0.988 & 0.013 \\
\bottomrule
\end{tabular}
\end{table}

Table~\ref{tab:exp2} reports both correlation levels. The real-calibrated panel reproduces the
morphology of the synthetic one: in the weak regime arm A's FDR is $18\%$--$42\%$ while arm B's
is $\approx0$, with abstention of $93\%$--$99\%$. The conclusion is therefore not an artifact of
the i.i.d.\ assumption. Holding $t=1.44$ fixed, $\Delta$FDR for $N\in\{20,40,80\}$ is $+0.250$,
$+0.200$, $+0.317$---all positive, so neither candidate count nor ground-truth share changes the
direction of the result.

\subsection{Diagnostic experiment: the excess-return requirement}
\label{sec:diag}

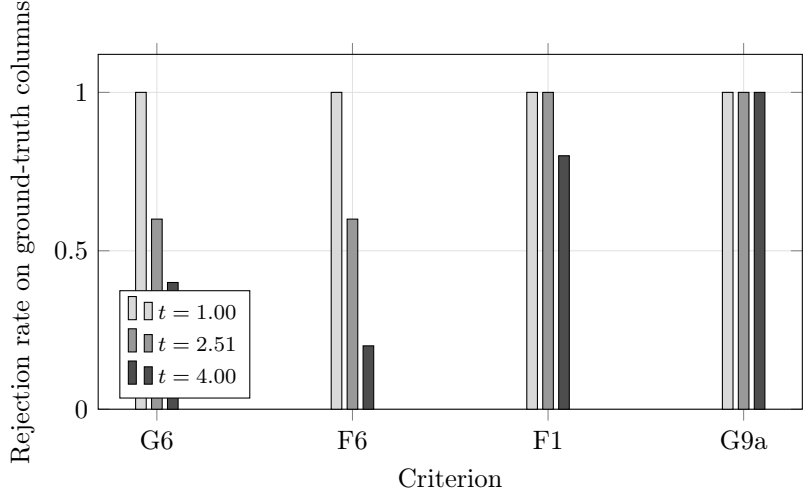
\begin{figure}[t]
\centering
\begin{tikzpicture}
\begin{axis}[
  width=0.66\textwidth, height=0.38\textwidth,
  ybar, bar width=4pt,
  xlabel={Criterion}, ylabel={Rejection rate on ground-truth columns},
  symbolic x coords={G6, F6, F1, G9a}, xtick=data,
  ymin=0, ymax=1.12, legend pos=south west, legend cell align=left,
  grid=major, grid style={line width=.2pt, draw=black!12},
  tick label style={font=\small}, label style={font=\small}, legend style={font=\scriptsize}]
\addplot[fill=black!15, draw=black] coordinates {(G6,1.00) (F6,1.00) (F1,1.00) (G9a,1.00)};
\addlegendentry{$t=1.00$}
\addplot[fill=black!40, draw=black] coordinates {(G6,0.60) (F6,0.60) (F1,1.00) (G9a,1.00)};
\addlegendentry{$t=2.51$}
\addplot[fill=black!70, draw=black] coordinates {(G6,0.40) (F6,0.20) (F1,0.80) (G9a,1.00)};
\addlegendentry{$t=4.00$}
\end{axis}
\end{tikzpicture}
\caption{Per-criterion rejection rate on \emph{ground-truth} columns before the excess-return
correction. The DSR rejects a true signal $100\%$ of the time at every strength: under absolute
returns the market component inflates volatility and compresses the effective $t$-statistic by
roughly a factor of three.}
\label{fig:reject}
\end{figure}

Before the correction of Section~\ref{sec:excess}, arm B abstained $100\%$ of the time even at
$t=4.0$. Figure~\ref{fig:reject} and Table~\ref{tab:diag} give per-criterion rejection rates for ground-truth columns and
identifies the cause: under absolute returns the market component dominates total volatility and
compresses the effective $t$-statistic by roughly a factor of three, so the DSR never passes.
After the correction, arm B at $t=4.0$ attains a hit rate of $98.3\%$ while keeping FDR at zero.

\begin{table}[htbp]
\centering
\caption{Per-criterion rejection rate for ground-truth columns (pre-correction), Panel II.}
\label{tab:diag}
\small
\begin{tabular}{@{}lrrr@{}}
\toprule
\textbf{Criterion} & $t=1.00$ & $t=2.51$ & $t=4.00$ \\
\midrule
G6 sub-period stability & 1.00 & 0.60 & 0.40 \\
F6 leave-one-trade-out & 1.00 & 0.60 & 0.20 \\
F1 sample size ($N^{*}$) & 1.00 & 1.00 & 0.80 \\
\textbf{G9a DSR} & $\mathbf{1.00}$ & $\mathbf{1.00}$ & $\mathbf{1.00}$ \\
\bottomrule
\end{tabular}
\end{table}

\noindent\textbf{Finding (central).} The failure is \emph{silent}: it presents as ``the framework
found nothing,'' easily misread as ``the candidates are poor.'' Any gate framework should define
criteria on excess returns and, at deployment, print per-criterion rejection rates for
ground-truth versus noise columns.

\subsection{Ablation: the sample-size criterion is redundant}
\label{sec:ablation}

Across the three Panel~I settings, arm B$'$ and arm B produce \emph{bit-identical} results: the
$N^{*}$ criterion is fully covered by the DSR in this regime. This does not establish that
$N^{*}$ is useless---when trials are few and single-run sample size binds, it may still be
necessary---but it is reported as a negative ablation result.

\subsection{Task morphology matters}
\label{sec:morphology}

Treating the 29 real ETFs as candidates (Panel~III), arm A attains a $100\%$ hit rate and FDR
$=0$ at \emph{every} signal strength, while gates only cause missed opportunities. The cause is
the $0.504$ mean cross-sectional correlation: assets receiving injected $\alpha$ systematically
outperform the rest, so the training-period winner is necessarily a ground-truth signal. An
injected-truth protocol must match the candidate structure of the task under study, or one may
wrongly conclude that gates are harmful.

\section{Discussion}
\label{sec:discussion}

\subsection{What gates are for}

Our results position admission gates as a \textbf{precision device}, not a return device. Use
them when the candidate count is large, the training window is limited, and the cost of a wrong
adoption exceeds the cost of a miss---for example when capital is at stake. Do not use them when
the signal is already strong (training $t>2.5$), where they only reduce the adoption rate.

The result resonates with the replication literature. \citet{ioannidis} notes that a finding is
less likely to be true when effect sizes are small and many relationships are probed---the regime
in which our gates help. Conversely \citet{harveyliuzhu} and \citet{houxuezhang} document that
failures concentrate in exactly the marginal cases our gates reject. Gates are therefore best
understood as implementing, at decision time, the multiple-testing discipline this literature
prescribes after the fact.

\subsection{Relation to prior work}

Section~\ref{sec:related} positions this paper; two points bear emphasis here. First, relative
to robustness scoring \citep{minerva,gtscore} we change the object of study from a score to a
decision rule, and the excess-return requirement (\S\ref{sec:excess}) applies to those scores as
much as to ours. Second, relative to agentic mining
\citep{factorMiner,quantaalpha,alphaagent,automate,alphaseek} our contribution is composable:
gates can serve as their admission layer, and improved search efficiency does not by itself make
admission more reliable.

\subsection{Limitations}

\begin{enumerate}\itemsep1pt
\item \textbf{Synthetic component.} Ground truth is injected as a constant effect; real alpha
      decay and regime dependence are more complex. Panel~II uses real return series as the common
      factor---preserving kurtosis $7.84$ and cross-correlation $0.504$---and reproduces the
      synthetic morphology, which mitigates but does not eliminate the concern.
\item \textbf{Single market, single asset class} (A-share ETFs, 29 assets / 745 points).
\item \textbf{Incomplete criterion set.} Difference-trade auditing, selection
      degree-of-freedom diagnostics, and PBO/SPA are implemented in the framework but not
      exercised here.
\item \textbf{Thresholds are not calibrated.} The DSR level of $0.95$ and the $2\sigma/\mu$
      coefficient follow prior choices; their sensitivity is not scanned.
\item \textbf{Abstention has no utility model.} Abstention is recorded as ``not adopted'' without
      an explicit trade-off between the cost of a miss and the cost of a wrong adoption.
\end{enumerate}

\subsection{Future work}

Introduce an explicit utility function to ground thresholds in decision theory; extend the
protocol to PBO/SPA and difference-trade auditing to map criterion redundancy; repeat across
markets and asset classes; and apply it to audit the admission stages of published methods.

\section{Conclusion}
\label{sec:conclusion}

We tested whether admission gates in quantitative research are effective, using an injected-truth
protocol on synthetic independent candidates and on strategy-family panels calibrated to a real ETF
panel. Gates carry genuine information---at equal adoption rates their false discovery rate is well
below that of random admission ($\Delta$FDR $0.10$--$0.60$), so they are not merely a stricter
threshold. But the value has a boundary: it materialises only for training $t<2.5$, at the cost of
an adoption rate of $1\%$--$7\%$; beyond $t\approx2.5$ pure search suffices and gating is pure
overhead. Criteria must be defined on excess returns, or they silently reject all candidates. A
sample-size criterion proved redundant here, and in an asset-selection morphology the gates were
unhelpful.

In one sentence: admission gates increase \emph{decisional precision} at a substantial cost in
\emph{recall}. Treating them as return enhancers is a category error; treating them as a
conservative decision device under uncertainty---and validating, with the control protocol above,
that they beat random admission at the same adoption rate---is their correct role.

\appendix

\section{Reproducibility}
\label{app:repro}

Every number in this paper is reproducible from the accompanying artefacts:

\begin{Verbatim}[fontsize=\small]
python3 exp1_injected_gate_efficacy.py --reps 400
python3 exp1_injected_gate_efficacy.py --reps 300 \
        --n-cand 20 --k-true 5 --alphas 0.0008
python3 exp2_real_based_validation.py --reps 400 \
        --panel-kind strategy --sigma-e 0.007 \
        --alphas 0.00039 0.00059 0.00098 0.00156
python3 exp2_real_based_validation.py --reps 400 \
        --panel-kind strategy --sigma-e 0.004 \
        --alphas 0.00022 0.00034 0.00056 0.00089
python3 exp2_real_based_validation.py --reps 300 \
        --panel-kind asset \
        --alphas 0.00116 0.00174 0.00289 0.00463
\end{Verbatim}

Data anchor: \texttt{etf\_panel\_ts.frozen-20260930.json} (29 ETFs, 746 days, backward-adjusted).
Results are written to \texttt{*\_results.json}.



\begin{thebibliography}{99}

\bibitem[Brock et al.(1992)]{brock1992}
W.~Brock, J.~Lakonishok, and B.~LeBaron.
\newblock Simple technical trading rules and the stochastic properties of stock returns.
\newblock \emph{The Journal of Finance}, 47(5):1731--1764, 1992.

\bibitem[Bailey and L\'opez de Prado(2014)]{dsr}
D.~H. Bailey and M.~L\'opez de Prado.
\newblock The deflated Sharpe ratio: Correcting for selection bias, backtest overfitting, and
non-normality.
\newblock \emph{The Journal of Portfolio Management}, 40(5):94--107, 2014.
\newblock DOI: \href{https://doi.org/10.3905/jpm.2014.40.5.094}{10.3905/jpm.2014.40.5.094}.

\bibitem[Bailey et al.(2016)]{pbo}
D.~H. Bailey, J.~M. Borwein, M.~L\'opez de Prado, and Q.~J. Zhu.
\newblock The probability of backtest overfitting.
\newblock \emph{Journal of Computational Finance}, 20(4), 2016.

\bibitem[Hansen(2005)]{hansen2005}
P.~R. Hansen.
\newblock A test for superior predictive ability.
\newblock \emph{Journal of Business \& Economic Statistics}, 23(4):365--380, 2005.

\bibitem[Harvey and Liu(2015)]{harveyliu}
C.~R. Harvey and Y.~Liu.
\newblock Backtesting.
\newblock \emph{The Journal of Portfolio Management}, 42(1), 2015.
\newblock (Page ranges in secondary sources differ, 12--28 versus 13--28; verify against the
publisher record.)

\bibitem[Harvey et al.(2016)]{harveyliuzhu}
C.~R. Harvey, Y.~Liu, and H.~Zhu.
\newblock \ldots\ and the cross-section of expected returns.
\newblock \emph{The Review of Financial Studies}, 29(1):5--68, 2016.
\newblock DOI: \href{https://doi.org/10.1093/rfs/hhv059}{10.1093/rfs/hhv059}.

\bibitem[Hou et al.(2020)]{houxuezhang}
K.~Hou, C.~Xue, and L.~Zhang.
\newblock Replicating anomalies.
\newblock \emph{The Review of Financial Studies}, 33(5):2019--2133, 2020.
\newblock DOI: \href{https://doi.org/10.1093/rfs/hhy131}{10.1093/rfs/hhy131}.

\bibitem[Ioannidis(2005)]{ioannidis}
J.~P.~A. Ioannidis.
\newblock Why most published research findings are false.
\newblock \emph{PLoS Medicine}, 2(8):e124, 2005.
\newblock DOI: \href{https://doi.org/10.1371/journal.pmed.0020124}{10.1371/journal.pmed.0020124}.

\bibitem[Lo and MacKinlay(1990)]{lomackinlay}
A.~W. Lo and A.~C. MacKinlay.
\newblock Data-snooping biases in tests of financial asset pricing models.
\newblock \emph{The Review of Financial Studies}, 3(3):431--467, 1990.

\bibitem[McLean and Pontiff(2016)]{mcleanpontiff}
R.~D. McLean and J.~Pontiff.
\newblock Does academic research destroy stock return predictability?
\newblock \emph{The Journal of Finance}, 71(1):5--32, 2016.
\newblock DOI: \href{https://doi.org/10.1111/jofi.12365}{10.1111/jofi.12365}.

\bibitem[Sullivan et al.(1999)]{stwbootstrap}
R.~Sullivan, A.~Timmermann, and H.~White.
\newblock Data-snooping, technical trading rule performance, and the bootstrap.
\newblock \emph{The Journal of Finance}, 54(5):1647--1691, 1999.
\newblock DOI: \href{https://doi.org/10.1111/0022-1082.00163}{10.1111/0022-1082.00163}.

\bibitem[White(2000)]{white2000}
H.~White.
\newblock A reality check for data snooping.
\newblock \emph{Econometrica}, 68(5):1097--1126, 2000.

\bibitem[Tang et al.(2025)]{alphaagent}
Z.~Tang, Z.~Chen, J.~Yang, J.~Mai, Y.~Zheng, K.~Wang, J.~Chen, and L.~Lin.
\newblock AlphaAgent: LLM-driven alpha mining with regularized exploration to counteract alpha
decay.
\newblock In \emph{Proc.\ 31st ACM SIGKDD Conf.\ (KDD '25)}, V.2, Toronto, Canada, August 2025.
\newblock arXiv:2502.16789; DOI: \href{https://doi.org/10.1145/3711896.3736838}{10.1145/3711896.3736838}.

\bibitem[Xiao et al.(2024)]{tradingagents}
Y.~Xiao, E.~Sun, D.~Luo, and W.~Wang.
\newblock TradingAgents: Multi-agents LLM financial trading framework.
\newblock \emph{arXiv preprint} arXiv:2412.20138, 2024.

\bibitem[Wang et al.(2026)]{factorMiner}
\newblock FactorMiner: A memory-augmented, self-evolving agent for alpha factor mining.
\newblock \emph{arXiv preprint} arXiv:2602.14670, 2026.

\bibitem[Han et al.(2026)]{quantaalpha}
J.~Han, S.~Zhang, W.~Li, and et al.
\newblock QuantaAlpha: An evolutionary framework for LLM-driven alpha mining.
\newblock \emph{arXiv preprint} arXiv:2602.07085, 2026.

\bibitem[Kou et al.(2024)]{automate}
Z.~Kou, H.~Yu, J.~Luo, and et al.
\newblock Automate strategy finding with LLM in quant investment.
\newblock \emph{arXiv preprint} arXiv:2409.06289, 2024.

\bibitem[MinervaScore(2026)]{minerva}
\newblock Equity strategy backtesting: Luck or edge? The MinervaScore as a statistical robustness
grade.
\newblock \emph{arXiv preprint} arXiv:2608.23808, 2026.

\bibitem[Sheppert(2026)]{gtscore}
A.~P. Sheppert.
\newblock The GT-Score: A robust objective function for reducing overfitting in data-driven trading
strategies.
\newblock \emph{arXiv preprint} arXiv:2602.00080, 2026.

\bibitem[Chen et al.(2026)]{rsisurvey}
M.~Chen, L.~Wang, and B.~Qu.
\newblock Recursive self-improvement in AI: From bounded self-refinement to autonomous research
loops.
\newblock \emph{arXiv preprint} arXiv:2607.07663, 2026.

\bibitem[Zhao et al.(2026)]{aide2}
B.~Zhao, Z.~Jiang, Y.~Wu, D.~Srikanth, and D.~Xu.
\newblock Recursive self-improvement of AI research agents.
\newblock \emph{arXiv preprint} arXiv:2609.26457, 2026.

\bibitem[AlphaSeek(2026)]{alphaseek}
\newblock AlphaSeek: Trajectory-level self-iterative factor mining framework for multi-source
financial data.
\newblock \emph{arXiv preprint} arXiv:2608.13913, 2026.

\end{thebibliography}
\end{document}